%% file: bare_conf.tex
\documentclass[conference]{IEEEtran}
\ifCLASSINFOpdf
\else
\fi

\usepackage{graphicx}
\usepackage{caption}

\usepackage{color}
\usepackage{float}
\usepackage{amsmath,amsfonts,amssymb,bm}

\usepackage{adjustbox}
\usepackage[noadjust]{cite}
\usepackage{pbox}
\usepackage{algorithm}
\usepackage[noend]{algpseudocode}
\usepackage{algorithm}
\usepackage[noend]{algpseudocode}
\usepackage{array}

\usepackage{nicefrac}
\usepackage{mathtools}

\usepackage{epstopdf}
\usepackage{xspace}
\usepackage{svg}

\makeatletter
\def\BState{\State\hskip-\ALG@thistlm}

\makeatother
\algnewcommand{\algorithmicgoto}{\textbf{go to}}%
\algnewcommand{\Goto}{\algorithmicgoto\xspace}%
\algnewcommand{\Label}{\State\unskip}

\usepackage[utf8]{inputenc}
\usepackage{tikz}
\usetikzlibrary{positioning}

\begin{document}
%
\title{Learning Multi-Modal Self-Awareness Models for Autonomous Vehicles from Human Driving}

\author{\IEEEauthorblockN{Mahdyar Ravanbakhsh\IEEEauthorrefmark{1},
	    Mohamad Baydoun\IEEEauthorrefmark{1}, 
		Damian Campo\IEEEauthorrefmark{1},
		Pablo Marin\IEEEauthorrefmark{2},
		David Martin\IEEEauthorrefmark{2},\\
		Lucio Marcenaro\IEEEauthorrefmark{1} and
		Carlo S. Regazzoni\IEEEauthorrefmark{1}}
	\IEEEauthorblockA{\IEEEauthorrefmark{1}DITEN,
		University of Genoa, Italy \\ Emails: \{mahdyar.ravan, mohamad.baydoun, damian.campo\}@ginevra.dibe.unige.it \\
		\{lucio.marcenaro, carlo.regazzoni\}@unige.it}
	\IEEEauthorblockA{\IEEEauthorrefmark{2}Carlos III University of Madrid, Spain.\\
		Email: \{dmgomez, pamarinp\}@ing.uc3m.es}}


%


\maketitle

\begin{abstract}
	This paper presents a novel approach for learning self-awareness models for autonomous vehicles. Proposed technique is based on the availability of synchronized multi-sensor dynamic data related to different maneuvering tasks performed by a human operator. It is shown that different machine learning approaches can be used to first learn single modality models using coupled Dynamic Bayesian Networks; such models are then correlated at event level to discover contextual multi-modal concepts. In the presented case, visual perception and localization are used as modalities. Cross-correlations among modalities in time is discovered from data and are described as probabilistic links connecting shared and private multi-modal DBNs at the event (discrete) level. Results are presented on experiments performed on an autonomous vehicle, highlighting potentiality of the proposed approach to allow anomaly detection and autonomous decision making based on learned self-awareness models. 
\end{abstract}


%
\IEEEpeerreviewmaketitle

\section{Introduction}
\label{sec:intro}
\input{sections/intro.tex}

\section{Multi-modal self-awareness models}
\label{sec:Method}
This section describes the two levels of self-awareness. The first is called private layer (PL), whereas the second is called shared level (SL) as has been proposed in \cite{icassp2018}. Each level is learned based on visual perception and localization, respectively. In addition,  a probabilistic framework based on switching dynamic system is used to learn the SL. An incremental hierarchy of Generative Adversarial Networks (GANs) is used to learn the PL model.
\subsection{Shared Level of self-awareness}
\label{subsec:SL}
\input{sections/SL.tex}
\subsection{Private Layer of self-awareness}
\label{subsec:PL}
\input{sections/PL.tex}

\section{Experimental Dataset}
\label{sec:datasets}
\input{sections/Dataset.tex}

\section{Experimental results}
\label{sec:resilys}
\input{sections/test_normal.tex}
\input{sections/test_abnormal.tex}

\section{Discussion}
\label{sec:Discussion}

\input{sections/discussion.tex}

\section{Conclusion}
\label{sec:Conclusion}
In order to improve the self-awareness model in this work we presented a multi-perspective approach to detect anomalies for moving agents. In the presented self-awareness model, tow levels are considered, shared level that we use a state-space representation from an external observer placed in the EC reference system. The private layer of self-awareness is constructed of a hierarchy of cross-modal GANs to learn a complex data distribution in a weakly-supervised manner. This model is used to represent the PL self-awareness model of autonomous embodied agents, where we face with a high diversity distribution. Scores of Discriminator network are used to approximate the complexity of data. Namely, a set of distance maps between prediction and the observation scores is used as a criteria for creating another level in the hierarchical structure. Such technique facilitates breaking and solving a complex data distribution in an incremental fashion. The experimental results on semi-autonomous ground vehicles show the capability of our methodology to recognize anomalies using multiple viewpoints, namely PL and SL. A future research path could consist in combining information from different sources for decision making and robust the proposed self-awareness model. In particular, situational awareness and self-reactions could be increased by modeling the cross-correlation between PL and SL by the multi-modal DBNs.



\bibliographystyle{IEEEtran}
\bibliography{ref}
%



\end{document}

%% file: sections/intro.tex
Self-awareness refers to system capability to recognize and predict its
own state, possible actions and the result of these actions on
the system itself and on its environment \cite{Schlatow2017}. Recent developments in signal processing and machine learning techniques can be useful to design autonomous systems equipped with a self-awareness module to make it possible to compare how much the current realization is similar to a previous experience of the same type while a given task is executed. The capability of predicting task evolution in normal conditions (\emph{i.e.}, when the task follows the rules learned in the previous experience) and jointly detecting abnormal situations that can rise based on such self-awareness is an important task that allows autonomous systems to increase their situational awareness and the effectiveness of the decision making sub-modules \cite{Campo2017,icassp2018}. Models of different self-awareness layers can be integrated in order to buildup a structured and multi-modal self-aware behavior for an agent.

In \cite{icassp2018} a self-awareness model was introduced that consists of two layers: Shared Level (SL) and Private Layer (PL). The analysis of observed moving agents for learning the models of normal/abnormal dynamics in a given scene from an external viewpoint, represents an emerging research field \cite{Bastani2016,Morris2008,Moll2010,wacv_plug,nabi2013temporal,lin2017tube,sebe2018abnormal,sabokrou2016fully,rabiee2016crowd}. The planned activity of an entity is one of such models that can be defined as the sequence of organized state changes (actions) that an entity has to perform in a specific context to achieve a task. This set of actions can be learned from examples and clustered into sequential discrete patterns of motions. The availability of a plan that associates the current state with an action class makes it possible to detect normal/abnormal situations in future repetitions of the same task. In general, computational models for abnormality detection are trained on a set of observations corresponding to standard behaviors. Accordingly, abnormalities can be defined as observations that do not match with the patterns previously learned as regular, \emph{i.e.}, behaviors that have not been observed before\cite{Ramik2014}.

An active self-aware action plan can be considered as a filter that makes it possible to predict and estimate state behaviors using linear and non linear dynamic and observation models. Switching Dynamical Systems (SDSs) are well-known Probabilistic Graphical Models (PGMs) that are capable to manage discrete and continuous dynamic variables in a jointly dynamical filter. Such systems have been used successfully to improve decision making and tracking capabilities \cite{Bastani2016}. In SDSs, each dynamical model for continuous state variable in successive time instants is associated with one of a discrete set of values of a random variable represented as an higher level motivation of that dynamic model. Externally shareable observations related to actions that are a function of agent position (plan) can be observed and active probabilistic plan models can be learned including hidden continuous and discrete states. The most used algorithm to take advantage of learned hierarchical probabilistic knowledge in the online phase is Markov Jump Linear Systems (MJLS) \cite{Doucet2000}. MJLS uses a combination of Kalman filter (KF) and particle filter (PF) to predict and update the continuous and discrete state space posterior probabilities.
In this paper, a MJLS is used to exploit learned knowledge for the SL that includes the capability to self-detect abnormality situations.

However, an agent can also infer dynamic models with respect to inner variables that it can observe from a First Person viewpoint while performing the same task; this private knowledge is only directly accessible to the agent itself. Detecting abnormalities by using a self-awareness model learned starting from private multisensorial first person data acquired by the agent can be possible, while doing the same task for which the self-awareness SL model has been obtained. Such a model can be defined as the Private Layer (PL) of self-awareness. An external observer has no access to such information, so not being able to directly detect PL abnormalities, while it can still evaluate SL abnormalities using third person models including shared variables. A PL model can allow an agent to be able to evaluate abnormalities related to PL and SL models, as it was shown in \cite{icassp2018}. However, previous works mostly rely on a high level of supervision to learn PL self-awareness models \cite{Olier2017a,Ramik2014}, while in this work, we propose a weakly-supervised method based on a hierarchy of Cross-modal Generative Adversarial Networks (GANs) \cite{NIPS2014_5423} for estimating PL models. Weakly supervised PL models not can also provide a level of information to boost the SL model as well as they can be used to provide a joint self aware multisensorial modality to cross-predict heterogeneous multimodal anomalies related to the same task execution. This paper describes a novel method to learn the PL model using an incremental hierarchy of GANs \cite{NIPS2014_5423,icip17}. The private camera views acquired by first-person images taken by an agent during task execution can be used together with related optical-flows to learn models using Generative Adversarial Networks (GANs), more appropriate for reducing high dimensionality for the visual modalities.

%% file: sections/SL.tex
This level of self-awareness focuses on the analysis of observed moving agents for understanding their dynamics in a given scene. This model is said to be shared as corresponding measurements can be directly observed by the agent itself or external observers.
\subsubsection{Learning of spatial activities}

\label{subsubsec:RSA}
Let $Z_K$ be the measured location of an object in a given reference system. A Kalman Filter (KF) based on an ``unmotivated model'' is used for tracking agents' motions. Such dynamic filter assumes that an observed object moves according to a random dynamic model described as: $X_{k+1} = AX_k + w_k$,
where $X_{k}$ represents the agent's generalized state composed of its coordinate positions and velocities at a time instant $k$, such that $X_k = [\boldsymbol{x} \hspace{0.2cm} \boldsymbol{\dot{x}}]^\intercal$. $\boldsymbol{x} \in \mathbb{R}^d$ and $\boldsymbol{\dot{x}} \in \mathbb{R}^d$. $d$ represents the dimensionality of the environment. $A = [A_1 \hspace{0.2cm} A_2]$ is a dynamic model matrix: $A_1 = [I_d \hspace{0.2cm}  0_{d,d}]^\intercal$ and $A_2 = 0_{2d,d}$. $I_n$ represents a square identity matrix of size $n$ and $0_{l,m}$ is a $l \times m$ null matrix. $w_k$ represents the prediction noise which is here assumed to be zero-mean Gaussian for all variables in $X_k$ with a covariance matrix $Q$, such that $w_k \sim \mathcal{N}(0,Q)$. The filter is called unmotivated 
and assumes that the agent remains fixed. In other words, it moves only due to random noisy fluctuations associated with $w_k$.
By considering a 2-dimensional reference system, i.e., $d=2$, the generalized state of an agent can be described as: $X_k = [x_k,y_k,\dot{x}_k,\dot{y}_k]^\intercal$.


Agents’ states $X_k$ produced by the unmotivated filter are used as an input for a Self Organizing Maps (SOM) that clusters similar information (quasi-constant velocities) based on a weighted distance. This clustering process prioritizes similar velocities (actions) by weighting more such component. Accordingly, the following distance function that uses the weights $\beta$ and $\alpha$ is used for training SOM, such that:

\begin{equation}\label{eq2}
{d}(\mathcal{X},\mathcal{Y}) =\sqrt{(\mathcal{X}-\mathcal{Y})^\intercal D(\mathcal{X}-\mathcal{Y})},
\end{equation} 
where $D = [\mathcal{B} \hspace{0.2cm} \mathcal{A}]$. $\mathcal{B} = [\beta I_2 \hspace{0.2cm} 0_{2,2}]^\intercal$, $\mathcal{A}  = [0_{2,2} \hspace{0.2cm} \alpha I_2]^\intercal$. $\mathcal{X}$ and $\mathcal{Y}$ are both 4-dimensional vectors of the form $[x \hspace{0.2cm} y \hspace{0.2cm} \dot{x} \hspace{0.2cm} \dot{y}]^\intercal$. 
The SOM’s output consists of a set of neurons encoding the main information from observed data inside a prototype structure that has the same form of the generalized states. Trained neurons represent a set of zones that segment the continuous state space, into a set of regions here called Superstates $S_k$. Each zone is composed of a set of clustered positions and their correspondent velocities which define quasi-constant velocity model used for tracking and prediction purposes.
Such quasi-constant velocity model can be defined as $[\dot{x}_n,\dot{y}_n]$, where $n$ indexes identifies the region (zone) corresponding to the superstates. We can define a linear dynamic model for each superstate as follows:

\begin{equation}\label{eq3}
X_{k+1} = FX_{k} + BU_{S_{k}} + w_k,
\end{equation}   

where $F = [F_1 \hspace{0.2cm} F_2]$ is a transition matrix that maps entities' positions as constant with respect to the previous state, such that $F_1 = [I_2 \hspace{0.2cm}  0_{2,2}]^\intercal$ and $F_2 =0_{4,2}$. $B = [I_2\Delta k \hspace{0.2cm} I_2]^\intercal $ is a control input model.  $U_{S_{k}} = [\dot{x}_k$, $\dot{y}_k]^T$, $k$ indexes the time, $\Delta k$ is the sampling time and $w_{k}$ is the process noise. The variable $U_{S_{k}}$ is a control vector that encodes the expected entity's velocity when its state falls in a discrete region $S_k$.


\subsubsection{ Online testing MJPF}
\label{subsubsec:MJM}
The SL model can be represented by means of a DBN switching model. Such model includes a discrete set of state regions subspaces corresponding to the vocabulary of switching variables. A set quasiconstant velocity model described before (Eq. \eqref{eq3}) is associated to one of such regions that describe a possible alternative relation between consecutive temporal states. A further learning step facilitates to obtain temporal transition matrices between superstates. This allows the system to estimate not only the next superstates but the moment where discrete transitions take place. Fig. \ref{fig:dbns}-a shows the proposed DBN used for modeling agents' behaviors, where arrows represent conditional probabilities: vertical arrows introduce causalities between both (discrete and continuous) levels of influence and observed measurements. Horizontal arrows explain temporal causalities between hidden variables. A Markov Jump Particle Filter (MJPF) is used to infer posterior probabilities on discrete and continuous states iteratively. MJPF essentially consists in a particle filter (PF) working at discrete level, embedding in each particle a Kalman filter (KF). Consequently, for each particle has attached a KF which depends on the superstate $S_K$ (see Eq. \eqref{eq3}). Such filter is used to obtain the prediction for the continuous state associated with a particle's superstate $S_k^*$, that is $p(X_k|X_{k-1},S_{k-1}^*)$; and the posterior probability $p(X_k,S_k^*|Z_{k})$ is estimated according to current observation $Z_k$. 

Abnormalities can be seen as deviations from predictions that the MJPF can do using the learned models embedded in its switching model  and new observed trajectories where interaction with the environment differ from training data.
Since a probabilistic filtering approach is considered, two main moments can be distinguished: $i)$ Prediction: which corresponds to an estimation of future states at a give time $k$. $ii)$ Update: computation of the generalized state posterior probability based on the comparison between predicted states and new measurements. Accordingly, abnormality behaviors can be measured in the update phase, i.e., when predicted probabilities are far from to observations. As it is well known, innovations in KFs are defined as:
\begin{equation}\label{eq4}
\epsilon_{k,n} = Z_k - H\hat{X}^n_{k|k-1},
\end{equation}
where $\epsilon_{k,n}$ is the innovation generated in the zone $n$ where the agent is located at a time $k$. $Z_k$ represents observed spatial data and $\hat{X}^n_{k|k-1}$ is the KF estimation of the agent's location at the future time $k$ calculated in the time instant $k-1$  Eq. \eqref{eq3}. Additionally, $H$ is the observation model that maps measurements into states, such that $Z_k = H X_k + v$ where $v \sim \mathcal{N}(0,R)$ and $R$ is the covariance observation noise.

Abnormalities can be seen as moments when a tracking system fails to predict subsequent observations, so that new models are necessary to explain new observed situations. A weighted norm of innovations is employed for detecting abnormalities, such that:
\begin{equation}\label{eq5}
\mathcal{Y}_k = \mathbf{d}(Z_k,H\hat{X}_{k|k-1}).
\end{equation}
In the MJPF, the expression shown \eqref{eq5} is computed for each particle and the median of such values is used as a global anomaly measurement of the filter. Further details about the implemented method can be found in \cite{fusion_valentina}.

%

\input{sections/dbn_sl.tex}

%% file: sections/dbn_sl.tex

\colorlet{rectangle edge}{blue!50}
\colorlet{rectangle area}{red!20}

\tikzset{filled/.style={fill=rectangle area, draw=rectangle edge, thick},
    outline/.style={draw=rectangle edge, thick}}

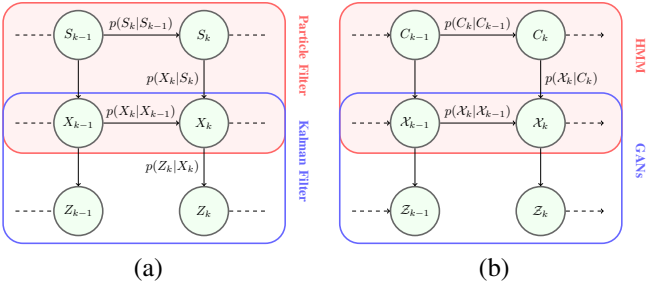
\begin{figure}
\begin{minipage}[t]{0.24\textwidth}
\centering
\scalebox{0.5}{
\begin{tikzpicture}[
roundnode/.style={circle, draw=black!60, fill=green!5, very thick, minimum size=1.3cm},
imaginarynode/.style={circle, very thick, minimum size=1mm},
]
\filldraw[color=red!60, fill=red!5, very thick, rounded corners=15pt](-2,-0.8) rectangle (5.5,3.2) node[rotate=-90]at (6,1.8) {\textbf{Particle Filter}};
\draw[color=blue!60, very thick, rounded corners=15pt](-2,0.8) rectangle (5.5,-3.2) node[rotate=-90]at (6,-1) {\textbf{Kalman Filter}};
\node[roundnode]        (xk_1)                              {$X_{k-1}$};
\node[roundnode]        (sk_1)                [above=of xk_1] {$S_{k-1}$};
\node[roundnode]        (zk_1)                [below=of xk_1] {$Z_{k-1}$};

\node[imaginarynode]    (sk_2)       [left=of sk_1] {};
\node[imaginarynode]    (xk_2)       [left=of xk_1] {};
\node[imaginarynode]    (zk_2)       [left=of zk_1] {};

\node[roundnode]    (xk)                [right=0cm and 2cm of xk_1] {$X_{k}$};
\node[roundnode]    (sk)                [above=of xk] {$S_{k}$};
\node[roundnode]    (zk)                [below=of xk] {$Z_{k}$};

\node[imaginarynode]    (sk_n)       [right=of sk] {};
\node[imaginarynode]    (xk_n)       [right=of xk] {};
\node[imaginarynode]    (zk_n)       [right=of zk] {};

\draw[dashed] (sk_2.east) -- (sk_1.west);
\draw[dashed] (xk_2.east) -- (xk_1.west);
\draw[dashed] (zk_2.east) -- (zk_1.west);

\draw[->] (sk_1.south) -- (xk_1.north);
\draw[->] (xk_1.south) -- (zk_1.north);

\draw[->] (xk_1.east) -- (xk.west) node[midway, above] {$p(X_k|X_{k-1})$};
\draw[->] (sk_1.east) -- (sk.west)  node[midway, above] {$p(S_k|S_{k-1})$};

\draw[->] (sk.south) -- (xk.north) node[midway, left] {$p(X_k|S_k)$};
\draw[->] (xk.south) -- (zk.north) node[midway, left] {$p(Z_k|X_k)$};

\draw[dashed] (sk.east) -- (sk_n.west);
\draw[dashed] (xk.east) -- (xk_n.west);
\draw[dashed] (zk.east) -- (zk_n.west);


\end{tikzpicture}
}
{(a)}
~
\end{minipage}
\begin{minipage}[t]{0.24\textwidth}
\centering
\scalebox{0.5}{
\begin{tikzpicture}[->,
roundnode/.style={circle, draw=black!60, fill=green!5, very thick, minimum size=1.3cm},
imaginarynode/.style={circle, very thick, minimum size=1mm},
rectanglenode/.style={rectangle, draw=black!60, fill=blue!5, very thick, minimum size=1cm},
]
\filldraw[color=red!60, fill=red!5, very thick, rounded corners=15pt](-2,-0.8) rectangle (5.5,3.2) node[rotate=-90]at (6,1.8) {\textbf{HMM}};
\draw[color=blue!60, very thick, rounded corners=15pt](-2,0.8) rectangle (5.5,-3.2) node[rotate=-90]at (6,-1) {\textbf{GANs}};
\node[roundnode]        (xk_1)                              {${\cal X}_{k-1}$};
\node[roundnode]        (sk_1)                [above=of xk_1] {$C_{k-1}$};
\node[roundnode]        (zk_1)                [below=of xk_1] {${\cal Z}_{k-1}$};

\node[imaginarynode]    (sk_2)       [left=of sk_1] {};
\node[imaginarynode]    (xk_2)       [left=of xk_1] {};
\node[imaginarynode]    (zk_2)       [left=of zk_1] {};

\node[roundnode]    (xk)                [right=0cm and 2cm of xk_1] {${\cal X}_{k}$};
\node[roundnode]    (sk)                [above=of xk] {$C_{k}$};
\node[roundnode]    (zk)                [below=of xk] {${\cal Z}_{k}$};

\node[imaginarynode]    (sk_n)       [right=of sk] {};
\node[imaginarynode]    (xk_n)       [right=of xk] {};
\node[imaginarynode]    (zk_n)       [right=of zk] {};

\draw[dashed] (sk_2.east) -- (sk_1.west);
\draw[dashed] (xk_2.east) -- (xk_1.west);
\draw[dashed] (zk_2.east) -- (zk_1.west);

\draw[->] (sk_1.south) -- (xk_1.north);
\draw[->] (xk_1.south) -- (zk_1.north);


\draw[->] (sk_1.east) -- (sk.west)  node[midway, above] {$p(C_k|C_{k-1})$};

\draw[->] (sk.south) -- (xk.north) node[midway, right] {$p({\cal X}_k|C_k)$};
\draw[->] (xk.south) -- (zk.north);
\draw[->] (xk_1.east) -- (xk.west) node[midway, above] {$p({\cal X}_k|{\cal X}_{k-1})$};;

\draw[dashed] (sk.east) -- (sk_n.west);
\draw[dashed] (xk.east) -- (xk_n.west);
\draw[dashed] (zk.east) -- (zk_n.west);

\end{tikzpicture}
}
{(b)}
\end{minipage}

\caption{Proposed DBN switching models for: (a) SL, (b) PL}
\label{fig:dbns}
\end{figure}

%% file: sections/PL.tex
To model the PL of self-awareness, a hierarchical structure of cross-modal GANs is employed. This set of cross-modal GANs \cite{NIPS2014_5423} is learned the normality by a sequence of observed images (${\cal I}$) synchronously collected with SL position data from a first person viewpoint paired with their corresponding optical-flow maps (${\cal O}$) as the direct observation of image changes consequent to joint agent and environment motion. 

In order to understand the relation between these two modalities, the hierarchy of cross-modal GANs is adopted and trained in a weakly-supervised manner. The only supervision here is provided by a subset of normal data related to a reference situation (corresponding to the unmotivated filter for the SL layer) to train the first level of the hierarchy that we called $Base GANs$. The $Base GANs$ provides a reference for the next levels of the hierarchy, in which all the further levels are trained in a self-supervised manner. The source of such self-supervision is the criteria provided by the $Base GANs$. The rest of this section is dedicated to explain the procedure of learning a single cross-modal GAN, constructing the hierarchy of GANs, and finally the online application of the learned model for prediction and anomaly detection.

\subsubsection{Learning the cross-modal representation}
\label{ssec:lgan}
GANs are generative deep networks and trained using only unsupervised data. The supervisory information in a GAN is indirectly provided by an adversarial game between two independent networks: a generator ($G$) and a discriminator ($D$). During training, $G$ generates new data and $D$ tries to distinguish whether its input is real (i.e., it is a training image) or it was generated by $G$. This competition between $G$ and $D$ is helpful in boosting the ability of both $G$ and $D$. To learn the normal pattern two channels are used as observations: appearance (i.e., raw-pixels) and motion (optical-flow images) for two cross-channel tasks. In the first task, optical-flow images are generated from the original frames, while in the second task appearance information is estimated from an optical flow image.
Specifically, let ${\cal I}_k$ be the $k$-th frame of a training video and ${\cal O}_k$ the optical-flow obtained using ${\cal I}_k$ and ${\cal I}_{k+1}$. ${\cal O}_k$ is computed using \cite{brox2004high}.
Two networks are trained: ${\cal N}^{{\cal I} \rightarrow {\cal O}}$, which is trained to generate optical-flow from frames (task 1) and ${\cal N}^{{\cal O} \rightarrow {\cal I}}$, which generates frames from optical-flow (task 2).
In both cases, inspired by \cite{DBLP:journals/corr/IsolaZZE16,icip17,sangineto2016self}, our architecture is composed by two fully-convolutional networks: the conditional generator $G$ and the conditional discriminator $D$. 
The $G$ network is the U-Net architecture \cite{DBLP:journals/corr/IsolaZZE16}, which is an encoder-decoder following with {\em skip connections} helping to preserve important local information. 
For $D$ the {\em PatchGAN} discriminator \cite{DBLP:journals/corr/IsolaZZE16,ravanbakhsh2017training} is proposed, which is based on a ``small'' fully-convolutional discriminator.

\noindent{\emph{Hierarchy of cross-modal GANs}:} 
As described in Sec.\ref{sec:intro}, the assumption is that the distribution of the normality patterns is under a high degree of diversity. In order to learn such distribution we suggest a hierarchical strategy for high-diversity areas by encoding the different distributions into the different hierarchical levels, in which, each subset of train data is used to train a different GAN. The process of partitioning subsets of data an approach similar to SL layer is used, implying the segmentation into regions of the information about the distance between the (preddicted) generated frames and optical flow images and the new acquired images and optical flow at successive time instant. To construct the proposed hierarchy of GANs, a recursive procedure is adopted. As shown in Alg. \ref{alg:hgan} the inputs of the procedure are represented by two sets: ${\cal Z}$ could be seen as the set of observation vectors which includes all the observations from the normal sequence of training data. Specifically, ${\cal Z}$ includes a set of coupled Frame-Motion maps, where ${\cal Z} = \{ [{\cal I}_k, {\cal O}_k] \}_{k=1,...,N}$, and $N$ is the number of total train samples. Besides, the input ${\cal V}_l$ is a subset of ${\cal Z}$, provided to train GANs for each individual level of the hierarchy. For instance, in case of the first level GANs, that acts as reference dynamic model, the initial set ${\cal V}_0$ is used to train two cross-modal networks ${\cal N}^{{\cal I} \rightarrow {\cal O}}_0$, and ${\cal N}^{{\cal O} \rightarrow {\cal I}}_0$. Note that, the only supervision here is the initial ${\cal V}_0$ to train the first level of the hierarchy, and the next levels are built accordingly using the supervision provided by the first level of GANs.
After training ${\cal N}^{{\cal I} \rightarrow {\cal O}}_0$, and ${\cal N}^{{\cal O} \rightarrow {\cal I}}_0$, we input $G^{{\cal I} \rightarrow {\cal O}}_0$ and $G^{{\cal O} \rightarrow {\cal I}}_0$ using each frame ${\cal I}$ of the entire set ${\cal Z}$ and its corresponding optical-flow image ${\cal O}$, respectively. The generators predict Frame-Motion couples as:
\begin{equation}\label{eq:pred_gan}
\begin{multlined}
\quad {\cal P} = \{ [{\cal P}^{\cal I}_k , {\cal P}^{\cal O}_k]\}_{k=1,...,N}\\ 
{\cal P}^{\cal I}_k = G^{{\cal O} \rightarrow {\cal I}}_0 ( {\cal O}_k),\quad
{\cal P}^{\cal O}_k = G^{{\cal I} \rightarrow {\cal O}}_0 ( {\cal I}_k)
\end{multlined}
\end{equation}
where ${\cal P}^{\cal I}_k$ and ${\cal P}^{\cal O}_k$ are $k$-th predicted image and predicted optical-flow, respectively.
The distance maps ${\cal X}$ between the observations $\cal Z$ and the predictions $\cal P$ for both channel are computed by the discriminators $D$:
\begin{equation}\label{eq:dist_gan}
\begin{multlined}
\quad \quad  {\cal X} = \{ [{\cal X}^{\cal I}_k , {\cal X}^{\cal O}_k]\}_{k=1,...,N}\\ 
\quad{\cal X}^{\cal I}_k = D^{{\cal O} \rightarrow {\cal I}}_0 ( {\cal I}_k,  {\cal O}_k) -  
D^{{\cal O} \rightarrow {\cal I}}_0 ( {\cal P}^{\cal I},  {\cal O}_k) ,\\\quad 
{\cal X}^{\cal O}_k = D^{{\cal I} \rightarrow {\cal O}}_0 ( {\cal O}_k,  {\cal I}_k) -  
D^{{\cal I} \rightarrow {\cal O}}_0 ( {\cal P}^{\cal O}_k,  {\cal I}_k)\quad 
\end{multlined}
\end{equation}
The distance maps ${\cal X}$ can be seen as the coupled image-motion innovation, where ${\cal X}^{\cal I}_k$ and ${\cal X}^{\cal O}_k$ are representing innovation related to the $k$-th portion of generalized state associated with image and optical-flow, respectively. The joint innovations $\{ [{\cal X}^{\cal I}_k , {\cal X}^{\cal O}_k]\}_{k=1,...,N}$ input to a self-organizing map (SOM) \cite{Kohonen2001} in order to cluster similar innovations on appearance-motion information. Similar to clustering position-velocity information in the shared layer, here the clustering is done to discretize the innovations (\emph{i.e.}, variations with respect to the reference GAN) on appearance-motion into a set of super-states. Specifically, the SOM's output is a set of neurons encoding the innovation information into a set of prototypes. Detected prototypes (clusters) provide the means of discretization for representing a set of super-states $\{{C_k}\}_{k=1,...,M_0}$, where $M_0$ is the number of detected clusters. 
Each of these cluster can also memorize input images and optical-flow instances that can be related with a given innovation, so reflecting couples position-velocity of SL layer.

It is expected that the clusters which are containing the training data should obtain lower score, since the innovation between the prediction and observation is lower on the associated image and optical-flow set. This is the criteria to detect the new distributions for learning new GANs, in which the clusters with high different self similar average scores can be considered as new distributions. The data sets attached to the new detected distributions build the new subsets ${\cal V}_{l}$ to train new networks ${\cal N}^{{\cal I} \rightarrow {\cal O}}_l$, and ${\cal N}^{{\cal O} \rightarrow {\cal I}}_l$ for the $l$-th level of the hierarchy, where $l=0,1,...,L$ is the level in the hierarchy. This procedure continues until no new distribution is detected. Then GANs and detected super-states in each level are stacked incrementally for constructing the entire hierarchical structure of GANs ${\cal H}_{l}$. Such incremental nature of the proposed method is similar to the one for learning different KFs for the SL layer, despite more general application of the GANs dynamic models and makes it a powerful model to learn a very complex distribution of data in a self-supervised manner.

\begin{algorithm}

\caption{Constructing the hierarchy of GANs}\label{euclid}
\label{alg:hgan}
\begin{algorithmic}[1]
\Require
\State $\theta:  \text{  Threshold parameter for train a new GAN}$
\State ${\cal Z}:$ \text{ Entire training sequences} ${{\cal Z} = \{ ({\cal I}_k, {\cal O}_k) \}_{k=1,...,N}}$ 
\State ${\cal V}_0: \text{  Subset of } {\cal Z}$
\State $l=0: \text{  Counter of hierarchy level }$
\Ensure
\State $[{\cal H}_l] \text{ Hierarchy of GANs}$
\Procedure{TRAINING HIERARCHY OF GANS}{}\label{marker}
\Label $\texttt{train:}$
\State $\text{Train networks } {\cal N}^{{\cal I}\rightarrow {\cal O}}, {\cal N}^{{\cal O} \rightarrow {\cal I}}\text{, with } {\cal V}_l$
\State $[{\cal H}_l] \gets \text{Trained networks }{\cal N}^{{\cal I}\rightarrow {\cal O}}, {\cal N}^{{\cal O} \rightarrow {\cal I}} $
\State ${\cal P} \gets {G} ({\cal Z}) \text{: predictions}$
\State ${\cal X} \gets ||{D} ({\cal Z}) - {D} ({\cal P})||_1 \text{: states}$
\State $\text{Clustering states: }SOM({\cal X}) \text{: super-states} \{C_k\}$
\For{$\text{each identified cluster}$}
\State $\mu \gets \text{Average score maps in this cluster}$ 
\If{$\mu \ge \theta $}
\State $ l= l +1 $
\State ${\cal V}_l \gets \text{Samples from this cluster in }{\cal Z}$
\State \Goto \texttt{train}
\EndIf
\EndFor
\State \textbf{return} $[{\cal H}_l]$
\EndProcedure
\end{algorithmic}
\end{algorithm}

In our experiments, we use a \emph{Hierarchy of GANs}, which consists of two levels: $Level0$ or \emph{base GAN} (reference), which is trained with a provided initialization subset ${\cal V}_0$ for representing the empty straight path (that plays here the role of the unmotivated filter for SL), and $Level1$ which is trained over the high-scored innovations clusters. In our case this clusters semantically represent curves as situations where the dynamic model describing changes in images and optical-flow differs from the one predicted by a model where a straight motion in a free space is assumed. Note that, selecting the straight empty path (in normal situation) as initialization subset is considered for the sake of simplicity, since it is the most efficient way to reach a target and so the variation with respect to this behavior could be described as differential actions with respect to going straight. Similarly, the discriminator's learned decision boundaries are also used to detect the abnormal events at testing time that is explained in the next section.

\subsubsection{Online testing GANs}
\label{sec:detection_gan}
Once the GANs hierarchy $\{{\cal H}_l\}$ is trained, it can be used for online prediction and anomaly detection tasks. Here we describe the testing phase for state/label estimation and detecting the possible abnormalities.

\noindent{\emph{Label estimation}: }At testing time we aim to estimate the state and detect the possible abnormality with respect to the training set. More specifically, input the test sample into the first level in the hierarchy of GANs, let ${D}^{{\cal I} \rightarrow {\cal O}}_0$ and ${D}^{{\cal O} \rightarrow {\cal I}}_0$ be the patch-based discriminators trained using the two channel-transformation tasks. Given a test frame ${\cal I}_k$ and its corresponding optical-flow image ${\cal O}_k$, we first produce the reconstructed ${\cal P}^{\cal O}_k$ and  ${\cal P}^{\cal I}_k$ using the first level generators ${G}^{{\cal I} \rightarrow {\cal O}}_0$ and ${G}^{{\cal O} \rightarrow {\cal I}}_0$, respectively. Then, the pairs of patch-based discriminators ${D}^{{\cal I} \rightarrow {\cal O}}$ and ${D}^{{\cal O} \rightarrow {\cal I}}$, are applied for the first and the second task, respectively. 
This operation results in two scores maps for the observation: $D^{{\cal I} \rightarrow {\cal O}}_0 ( {\cal O}_k,  {\cal I}_k)$ and $D^{{\cal O} \rightarrow {\cal I}}_0 ( {\cal I}_k,  {\cal O}_k)$, and two score maps for the prediction (the reconstructed data): $D^{{\cal I} \rightarrow {\cal O}}_0 ( {\cal P}^{\cal O}_k,  {\cal I}_k)$ and $D^{{\cal O} \rightarrow {\cal I}}_0 ( {\cal P}^{\cal I}_k,  {\cal O}_k)$. In order to estimate the state, we used Eq. \ref{eq:dist_gan} to generate the joint representation ${\cal X}_k = [{\cal X}^{\cal I}_k , {\cal X}^{\cal O}_k]$, where:
\begin{equation}\label{eq:dist_gan}
\begin{multlined}
\quad{\cal X}^{\cal I}_k = D^{{\cal O} \rightarrow {\cal I}}_0 ( {\cal I}_k,  {\cal O}_k) -  
D^{{\cal O} \rightarrow {\cal I}}_0 ( {\cal P}^{\cal I},  {\cal O}_k) ,\\
{\cal X}^{\cal O}_k = D^{{\cal I} \rightarrow {\cal O}}_0 ( {\cal O}_k,  {\cal I}_k) -  
D^{{\cal I} \rightarrow {\cal O}}_0 ( {\cal P}^{\cal O}_k,  {\cal I}_k)\quad 
\end{multlined}
\end{equation}
Accordingly, to estimate the current super-state we use innovation ${\cal X}_k$ with respect to the empty straight motion GAN to find closest SOM's detected prototypes. This procedure repeat for all the $l=0,1,..,L$ levels in the hierarchy $[{\cal H}_l]$.
This discrete situation estimation can be seen as a way to explore a switching model (see Fig. \ref{fig:dbns}-b), where in continues levels a hierarchy of GANs associated with different neurons are estimating the states and the discrete level can be modeled by an HMM working on neurons classifying innovations and their time transitions.

\noindent{\emph{Anomaly detection}: }Note that, a possible abnormality in the observation (e.g., an unusual object or an unusual movement) corresponds to an outlier with respect to the data distribution learned by ${\cal N}^{{\cal I} \rightarrow {\cal O}}_l$ and ${\cal N}^{{\cal O} \rightarrow {\cal I}}_l$ during training. The presence of the anomaly, results in a low value of $D^{{\cal O} \rightarrow {\cal I}}_l ( {\cal P}^{\cal I},  {\cal O}_k)$ and $D^{{\cal I} \rightarrow {\cal O}}_l ( {\cal P}^{\cal O}_k,  {\cal I}_k)$ (predictions), but a high value of $D^{{\cal O} \rightarrow {\cal I}}_l ( {\cal I}_k,  {\cal O}_k)$ and $D^{{\cal I} \rightarrow {\cal O}}_l ( {\cal O}_k,  {\cal I}_k)$ (observation). 
Hence, in order to decide whether an observation is normal or abnormal with respect to the scores from the current hierarchy level of GANs, we simply calculate the average value of the innovations between prediction and observation maps for both modalities, which is obtained from:
\begin{equation}\label{eq:dist}
\tilde{Y}_k = \overline{{\cal X}^{\cal I}_k} + \overline{{\cal X}^{\cal O}_k}
\end{equation}

The final representation of private layer for an observation ${\cal Z}_k = ({\cal I}_k , {\cal O}_k )$ consists of the computed $\tilde{Y}_k$ and estimated super-state $C_k$. We defined an error threshold $\tilde{Y}_{th}$ to detect the abnormal events: when all the levels in the hierarchy of GANs tag the sample as abnormal (e.g., dummy super-state) and the measurement $\tilde{Y}$ is higher than this threshold, current measurement is considered as an abnormality. Note that, the process is aligned closely to the one followed with SL layer, despite GAN are more powerful as they allow to deal with strong multidimensional inputs as well as with not linear dynamic models at the continuous level. This complexity is required by video variables involved in PL as different from low dimensional positional variable involved in SL.

%% file: sections/Dataset.tex
\begin{figure}[t]
	\vspace{-0.25cm}
	\centering
	\begin{minipage}[t]{0.25\textwidth}
		\centering
		\includegraphics[width=3.75cm, height=2.5cm]{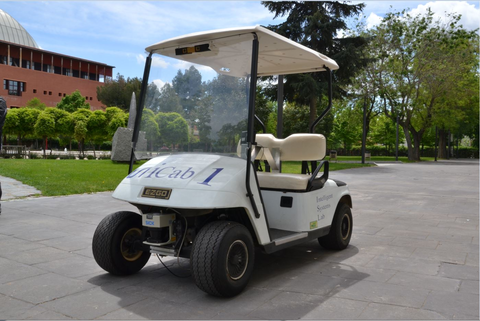}
		{Autonomous vehicle ``iCab''}
		
	\end{minipage}%
	~ 
	\begin{minipage}[t]{0.25\textwidth}
		\centering
		\includegraphics[width=3.75cm, height=2.5cm]{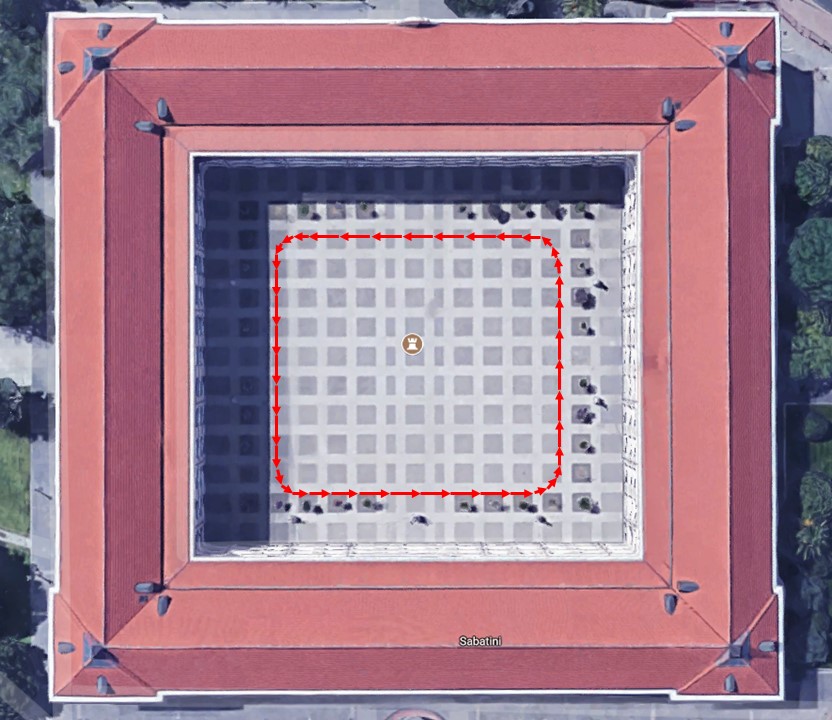}
		Infrastructure
		\label{fig:Environment}
	\end{minipage}
	\caption{Proposed moving entity and closed scene}
	\label{fig:iCab}
\end{figure}

In our experiments an iCab vehicle \cite{Marin2016} drove by a human operator is used to collect the dataset (see Fig. \ref{fig:iCab});  we obtained the vehicle's position mapped into Cartesian coordinates from the odometry manager~\cite{Marin2016}, as well as captured video footage from a first person vision acquired with a built-in camera of the vehicle. 
The observations are generated by taking state space position of iCab and estimating its flow components over the time. Furthermore, for the cross-modal GANs in the PL self-awareness model, we input the captured video frames and their corresponding optical-flow maps.

\begin{figure*}[t]
    \begin{center}
    \begin{minipage}[t]{0.32\textwidth}
		\centering
		\includegraphics[width=\textwidth]{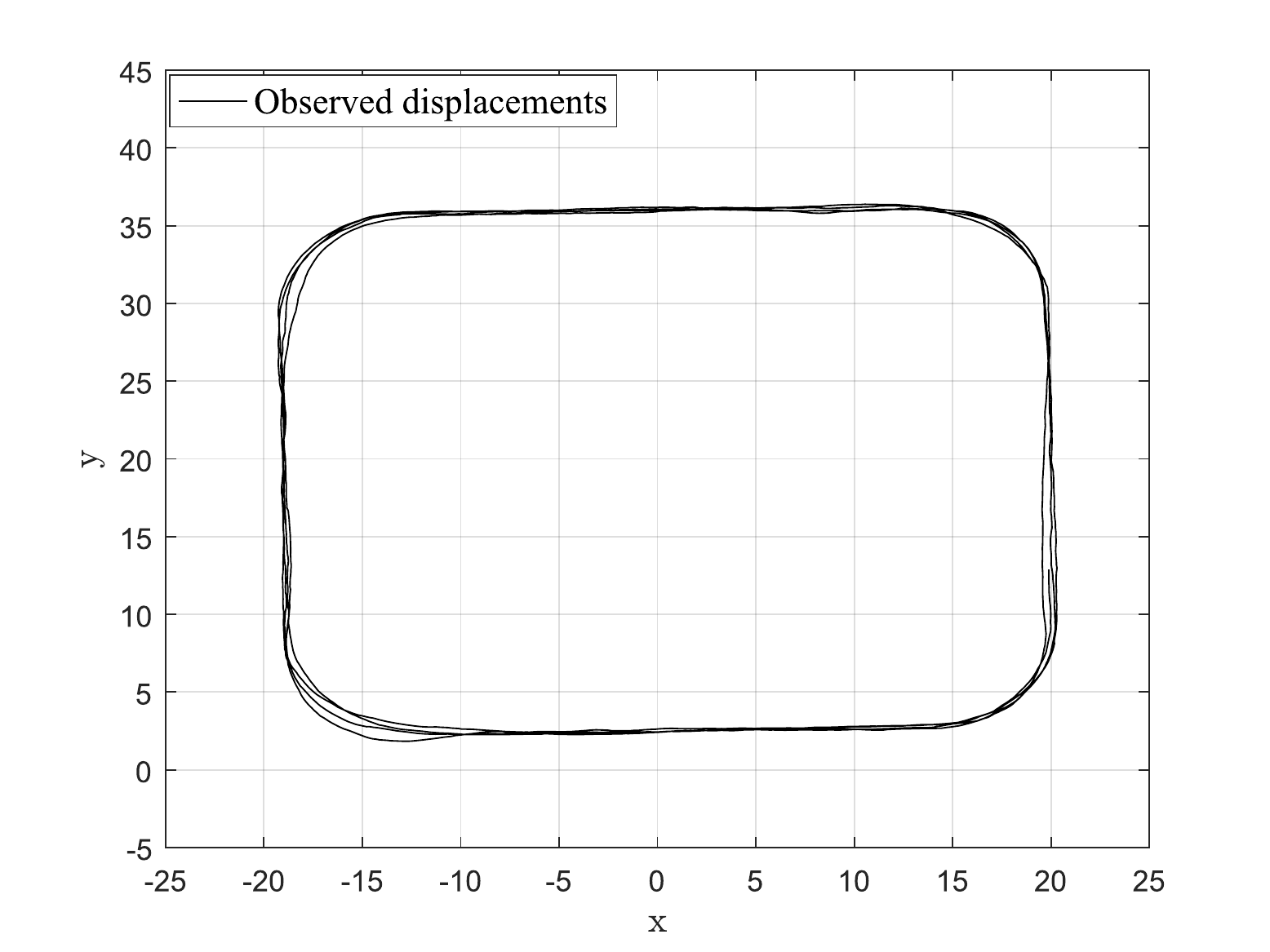}
		(a)
	\end{minipage}
	\begin{minipage}[t]{0.32\textwidth}
		\centering
		\includegraphics[width=\textwidth]{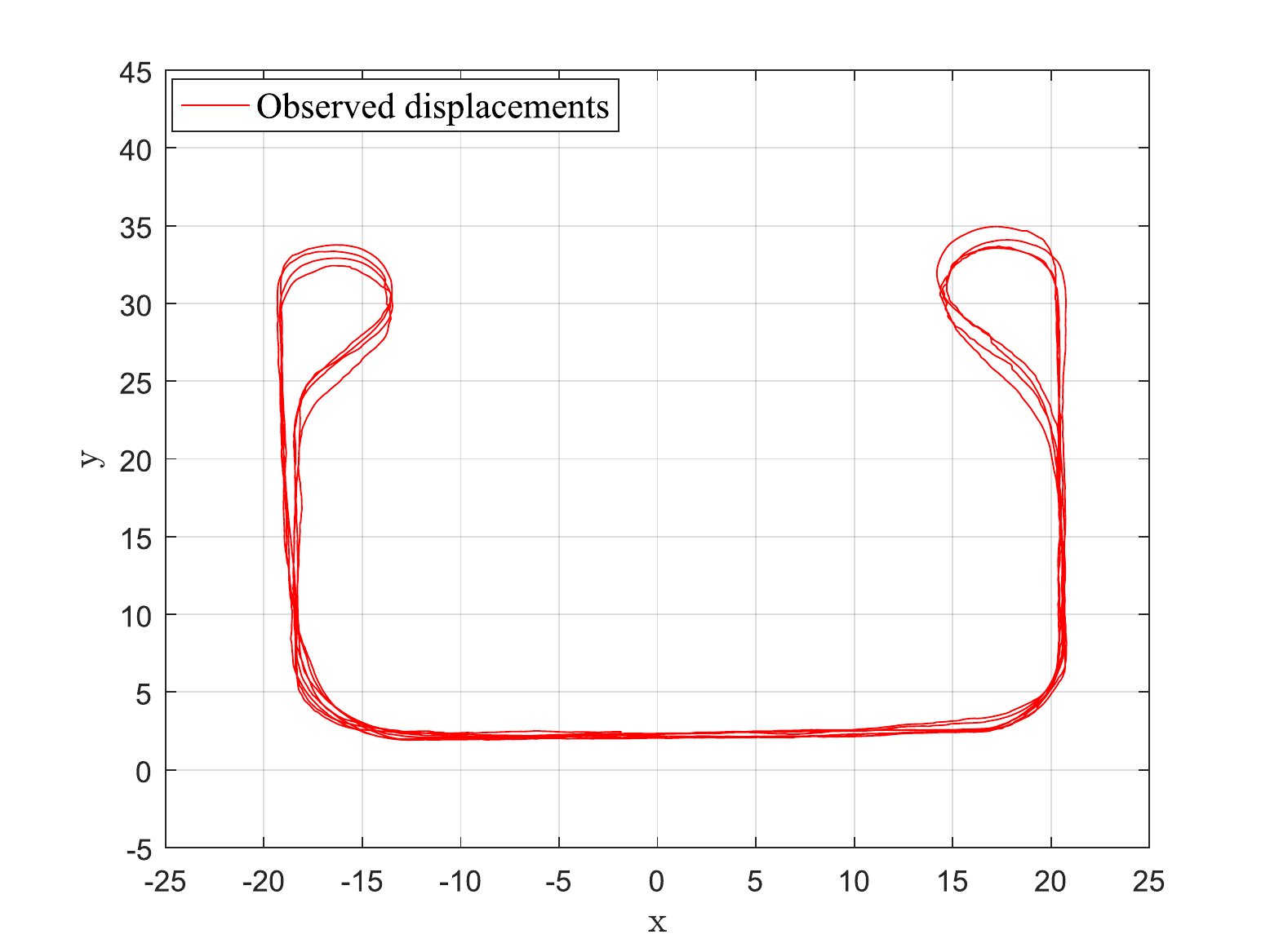}
		(b)
	\end{minipage}
	\begin{minipage}[t]{0.32\textwidth}
		\centering
		\includegraphics[width=\textwidth]{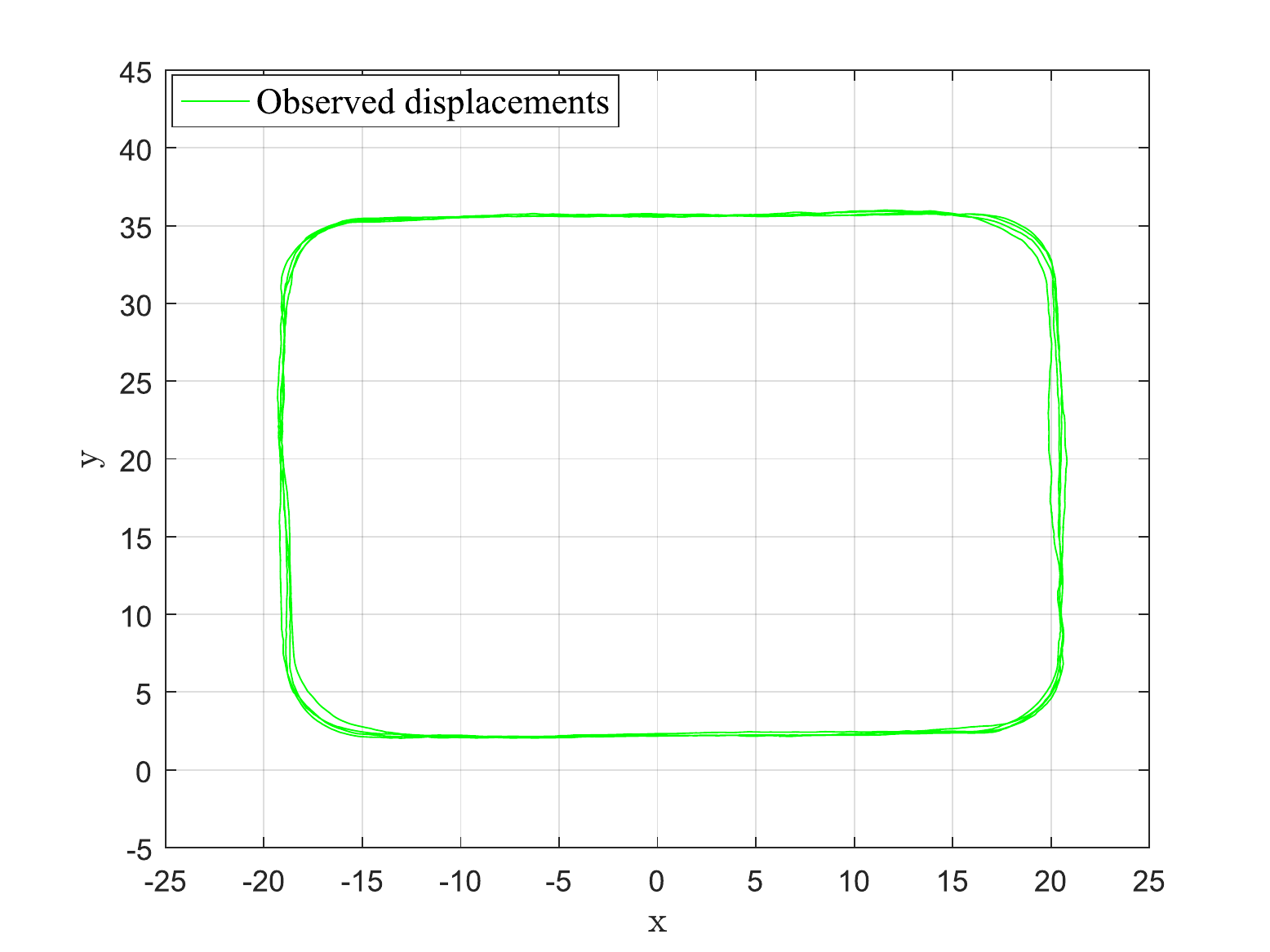}
		(c)
		\label{fig:Stop}
	\end{minipage}
	\end{center}
	\caption{Three different action scenarios: (a) perimeter monitoring under the normal situation (training set), and performing perimeter monitoring task in presence of abnormality (test sets): (b) U-turn, and (c) emergency stop.}
	\label{fig:plans}
\end{figure*} 
We aim to detect dynamics that have not been seen previously based on the normal situation (Scenario I) learned with the proposed method. Scenarios II and III includes unseen manoeuvres caused by the presence of pedestrians while the vehicle performs a perimeter control task. 
Accordingly, 3 situations (experiments) are considered in this work: \emph{Scenario I)} or normal perimeter monitoring, where the vehicle follows a rectangular trajectory around a building (see Fig.~\ref{fig:plans}-a). \emph{Scenario II)} or U-turn, where the vehicle performs a perimeter monitoring and is faced with a pedestrian, so it makes a U-turn to continue the task in the opposite direction (see Fig.~\ref{fig:plans}-b). \emph{Scenario III)} or emergency stop, where the vehicle encounters with pedestrians crossing its path and needs to stop until the pedestrian leaves its field of view (see Fig.~\ref{fig:plans}-c).


Situations II and III can be seen as deviations of the perimeter monitoring dynamics. 
 When an observation falls outside the superstate, as the learned model are not applicable, a {\em  dummy neuron} is used to represent the unavailability of an action from the learned experience and random filter where $U = 0_{2,1}$ in Eq.~\eqref{eq3} is considered for prediction to represent the uncertainty over state derivatives.

%% file: sections/test_normal.tex
\begin{figure}[t]
\centerline{\scriptsize{(a) }\includegraphics[width=0.96\linewidth,trim={4.4cm 0.6cm 3.2cm 0},clip]{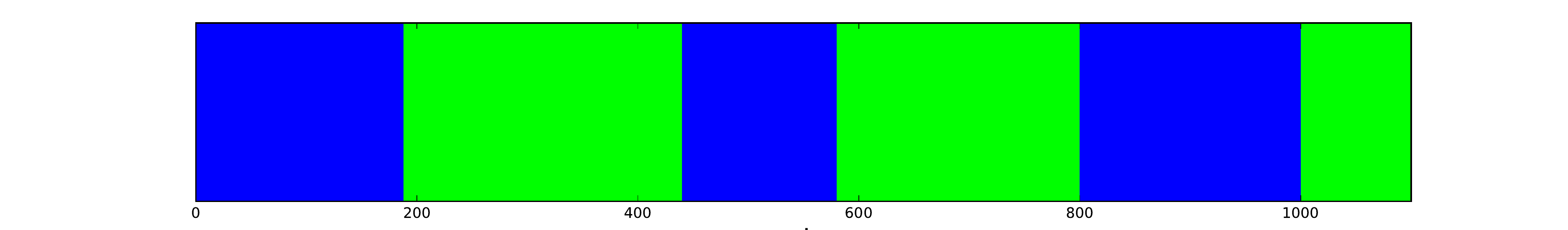}}
\centerline{\scriptsize{(b) }\includegraphics[width=0.96\linewidth,trim={4.4cm 0 3.2cm 0},clip]{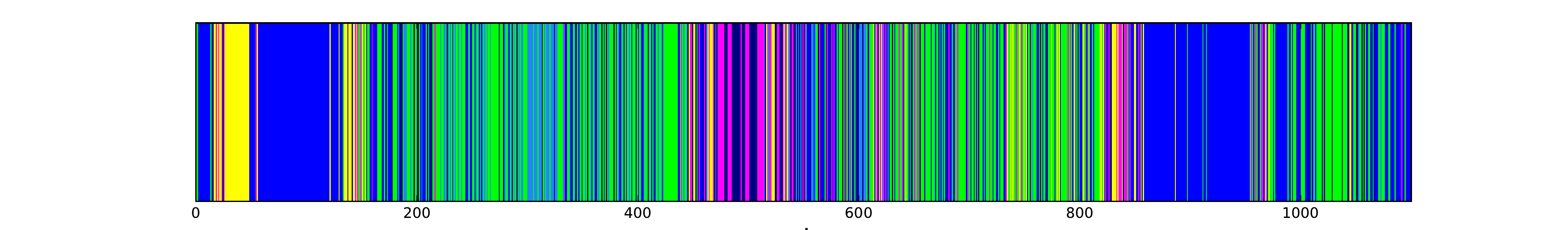}}
\centerline{\scriptsize{(c)  }\includegraphics[width=0.94\linewidth,trim={6cm 0 3.8cm 0},clip]{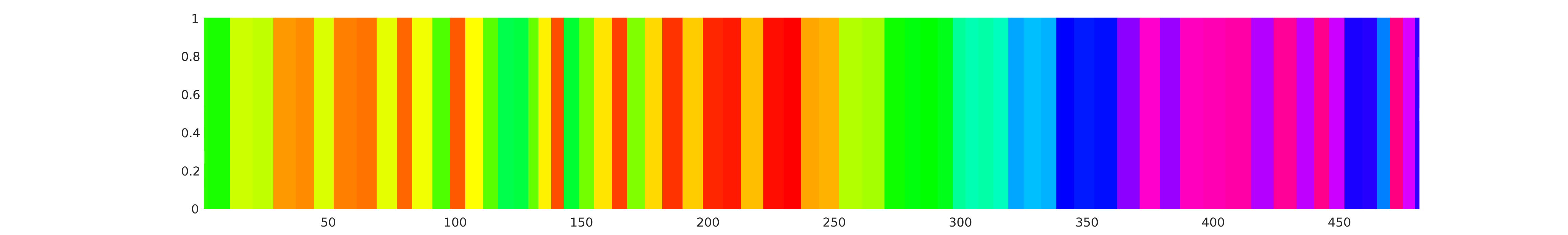}}
\centerline{\scriptsize{(d)}\includegraphics[width=0.96\linewidth,trim={4cm 0 3.2cm 0},clip]{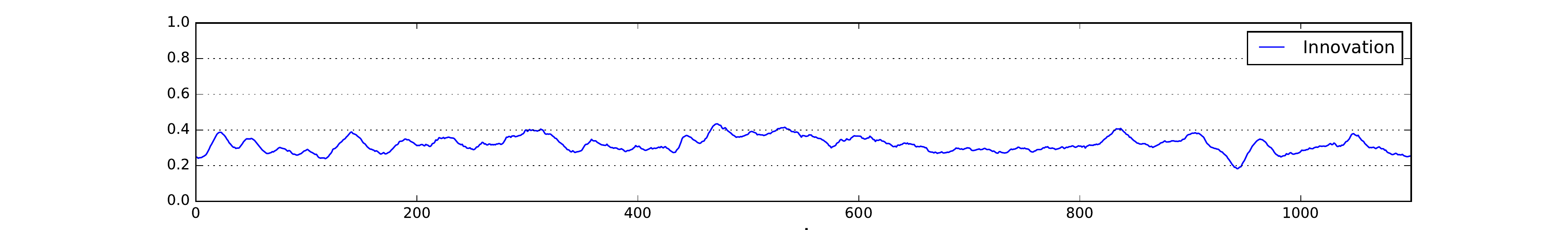}}
\centerline{\scriptsize{(e)}\includegraphics[width=0.96\linewidth,trim={4.8cm 0 3.5cm 0},clip]{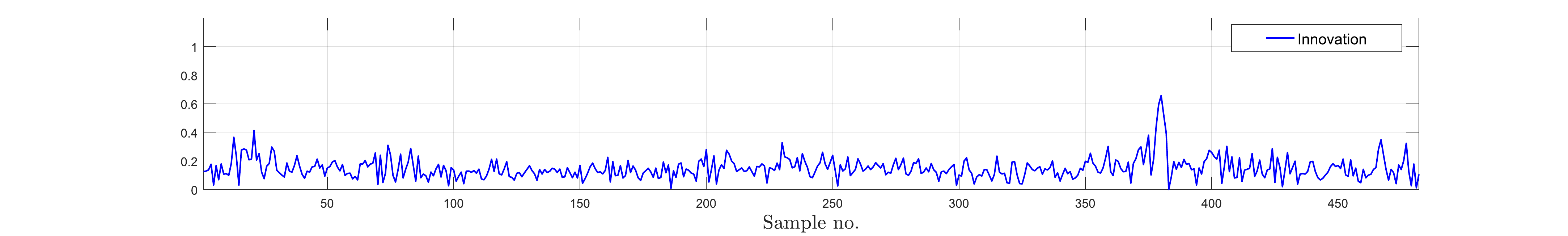}}
\caption{Normality representations from PL and SL: in (a) the ground truth labels are shown, moving straight is green and blue bars represent the curving. Color-coded super-states sequences $\{C_k\}$ and $\{S_k\}$ are shown in (b) and (c), respectively. They are highly correlated with the agent's real status (a). (d) and (e) show the abnormality signals from PL and SL, respectively. The horizontal axis represents the sample number, and the vertical axis shows the innovation values (abnormality signal).}
\label{fig:pri_states}
\end{figure}

\noindent{\textbf{Representation of normality}:} The two levels of the proposed self-awareness model, including the shared layer (modeled by MJPF), and the private layer (modeled as a hierarchy of GANs), are able to learn the normality. In our experiments this is defined as Scenario I (Fig.~\ref{fig:plans}-a) and it is used to learn both models. As reviewed in Sec. \ref{sec:Method} both SL and PL, represent their situation awareness by a set of super states following with abnormality signals. We select a period of normal perimeter monitoring task (see Fig. \ref{fig:subplans}-a) as a test scenario. The result for PL and SL is shown in Fig. \ref{fig:pri_states}, which simply visualizes the learned normality representations. The ground truth label is shown in Fig. \ref{fig:pri_states}-a, and the color-coded detected super states from PL $\{{C_t}\}$ and SL $\{S_k\}$ are illustrated in Fig. \ref{fig:pri_states}-b and Fig. \ref{fig:pri_states}-c, respectively. It clearly shows not only the pattern of superstates are repetitive and highly-correlated with the ground truth, but also there is a strong correlation between the sequence of PL and SL super states.

Note that the abnormality signals are stable for both PL and SL, while in case of an abnormal situation we expect to observe identical spikes over the signals. In order to study such abnormal situations we apply the trained self-awareness model over unseen test sequences. Two different scenarios are selected, in which the moving vehicle performing the perimeter monitoring task has to face abnormal events. In each scenario the agent performs different actions in order to solve the abnormal situation. The goal of this set of experiments is to evaluate the performance of the proposed self-awareness model (SL and PL) for detecting the abnormalities.

%% file: sections/test_abnormal.tex
\begin{figure*}[t]
    \begin{center}
    \begin{minipage}[t]{0.325\textwidth}
		\centering
		\includegraphics[width=\textwidth,trim={0.89cm 0.18cm 1.25cm 0.68cm},clip]{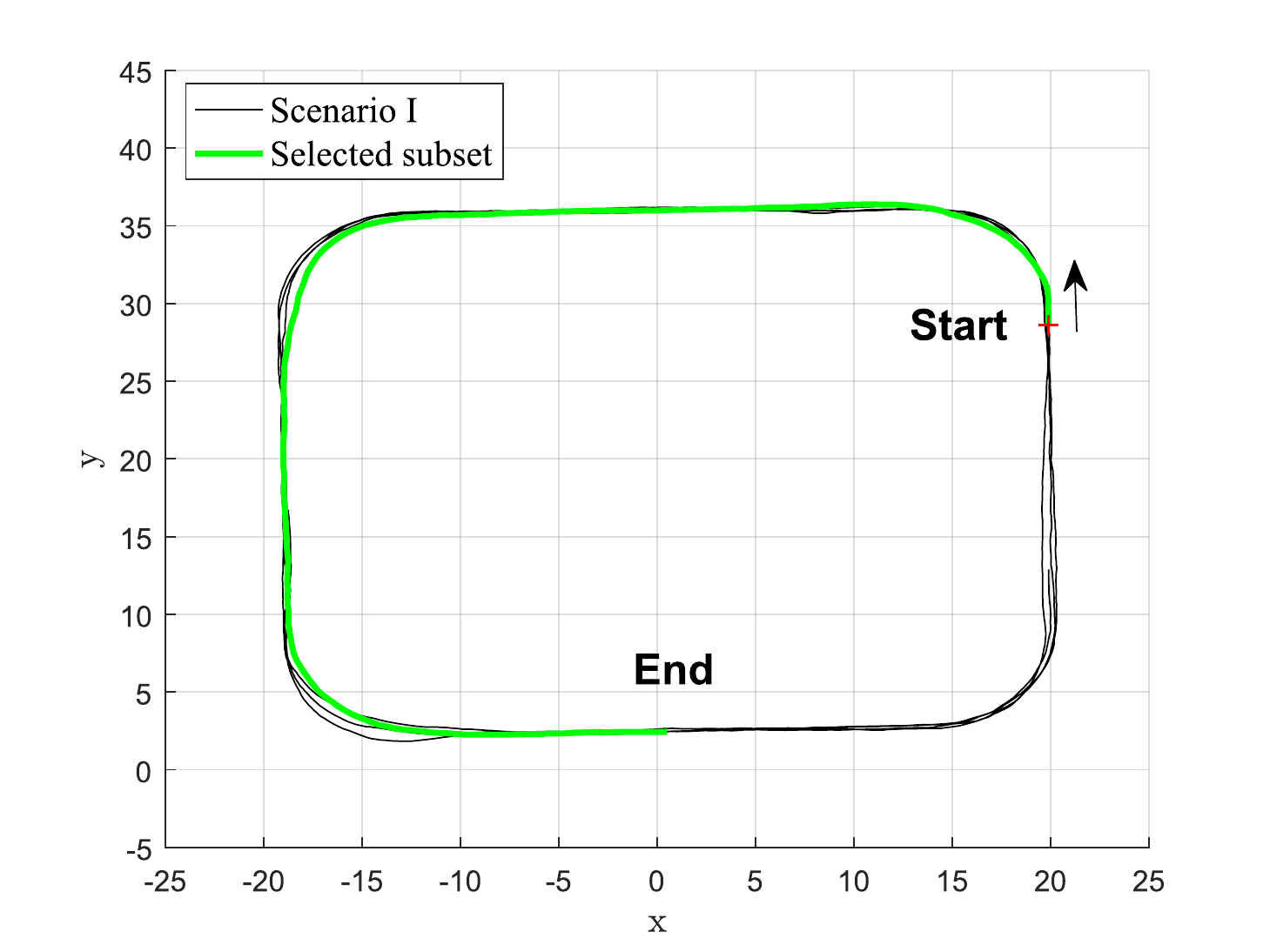}
		\scriptsize{(a) perimeter monitoring}
	\end{minipage}
	\begin{minipage}[t]{0.325\textwidth}
		\centering
		\includegraphics[width=\textwidth,trim={0.9cm 0.2cm 1.25cm 0.7cm},clip]{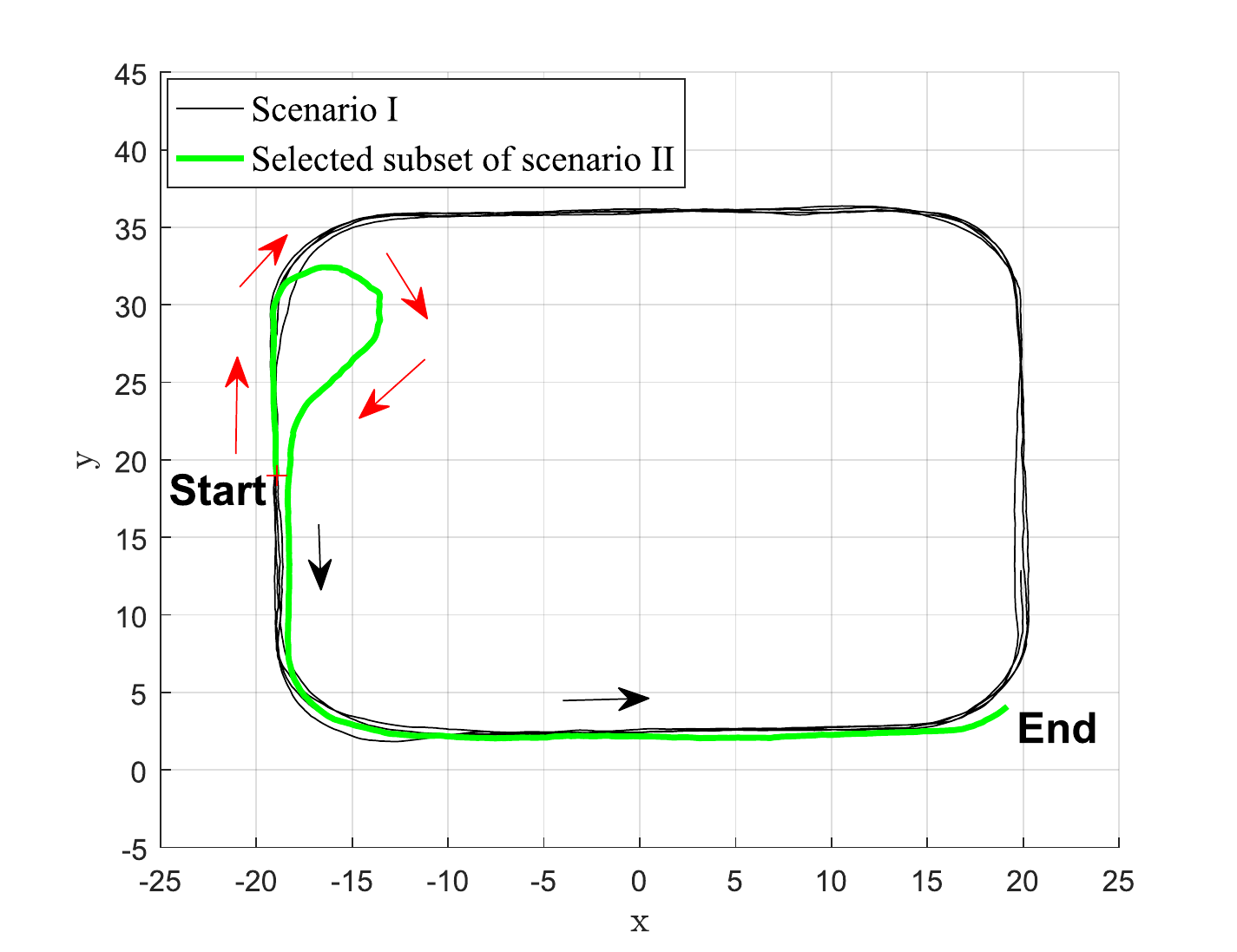}
		\scriptsize{(b) U-turn}
	\end{minipage}
	\begin{minipage}[t]{0.325\textwidth}
		\centering
		\includegraphics[width=\textwidth,trim={0.9cm 0.2cm 1.25cm 0.7cm},clip]{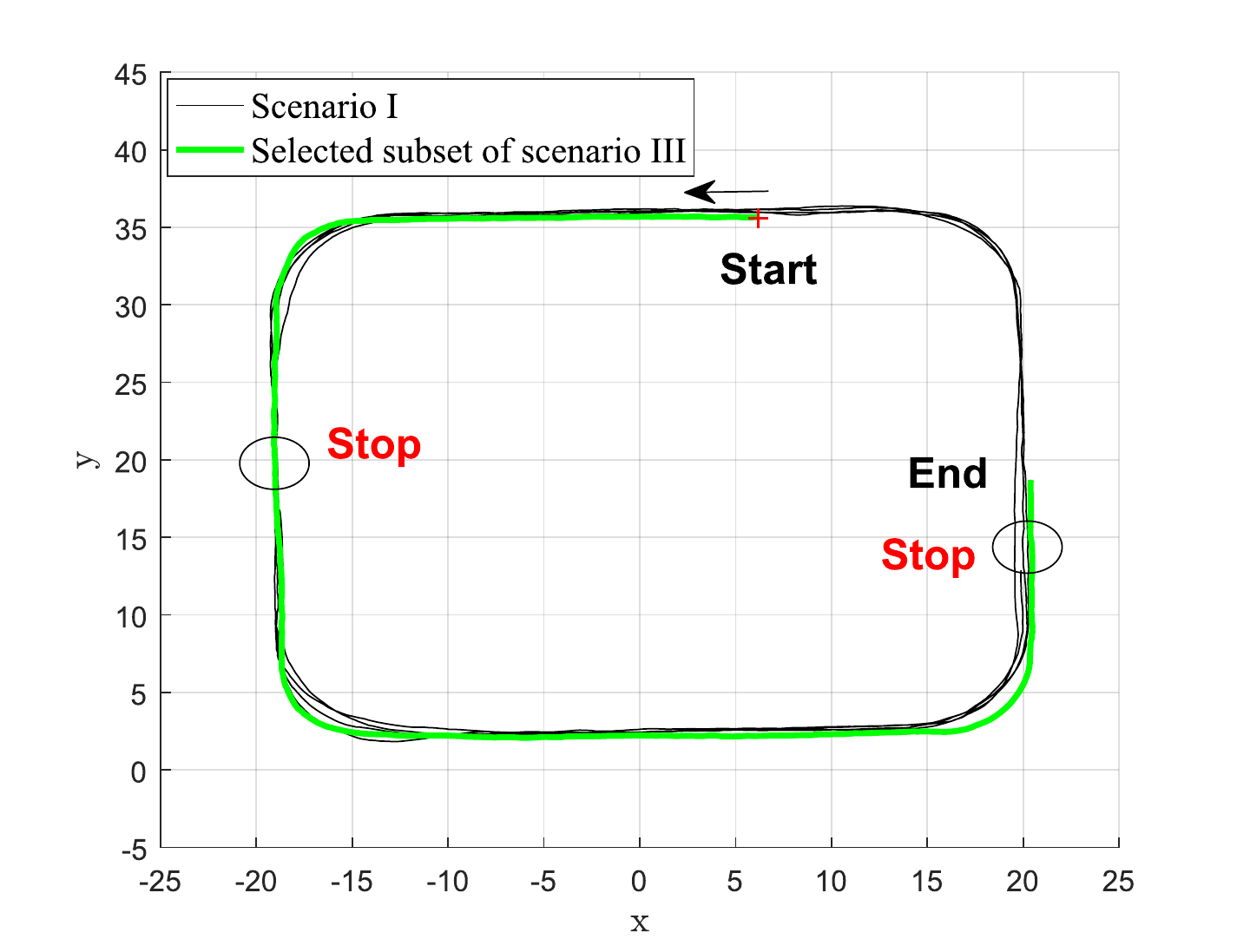}
		\scriptsize{(c) emergency stop}
		\label{fig:Stop}
	\end{minipage}
	\end{center}
	\caption{Sub-sequence examples from testing scenarios reported in the experimental results.}
	\label{fig:subplans}
\end{figure*} 

\begin{figure}[htb]
\centerline{\scriptsize{(a)}\includegraphics[width=0.96\linewidth,trim={4.35cm 0.6cm 3.2cm 0},clip]{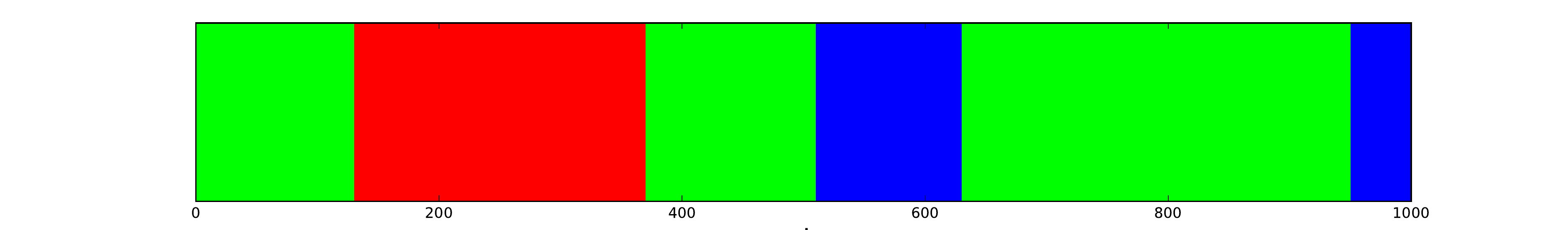}}
\centerline{\scriptsize{(b)}\includegraphics[width=0.96\linewidth,trim={4.35cm 0 3.2cm 0},clip]{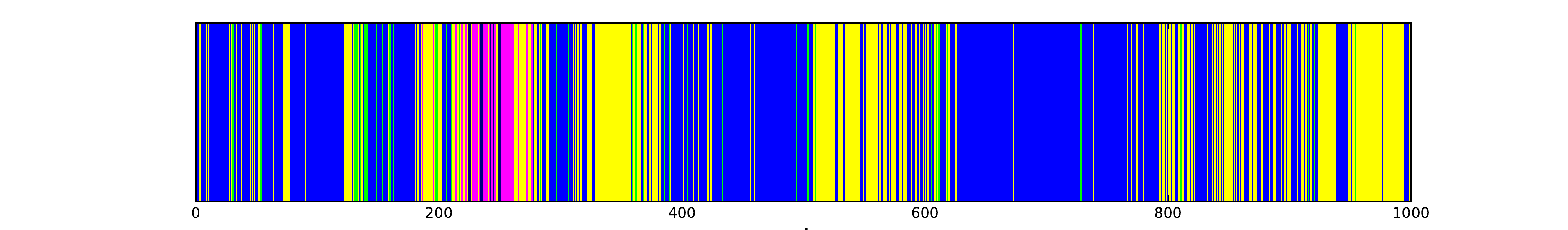}}
\centerline{\scriptsize{(c) }\includegraphics[width=0.94\linewidth,trim={6cm 0 3.8cm 0},clip]{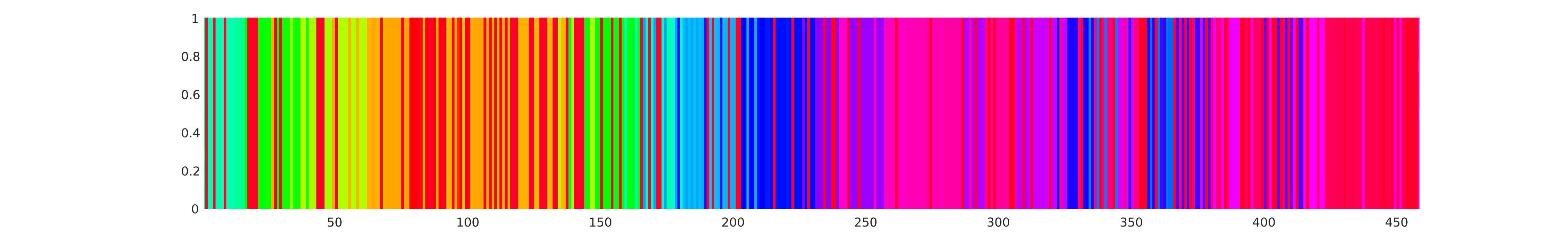}}
\centerline{\scriptsize{(d)}\includegraphics[width=0.96\linewidth,trim={4cm 0 3.2cm 0},clip]{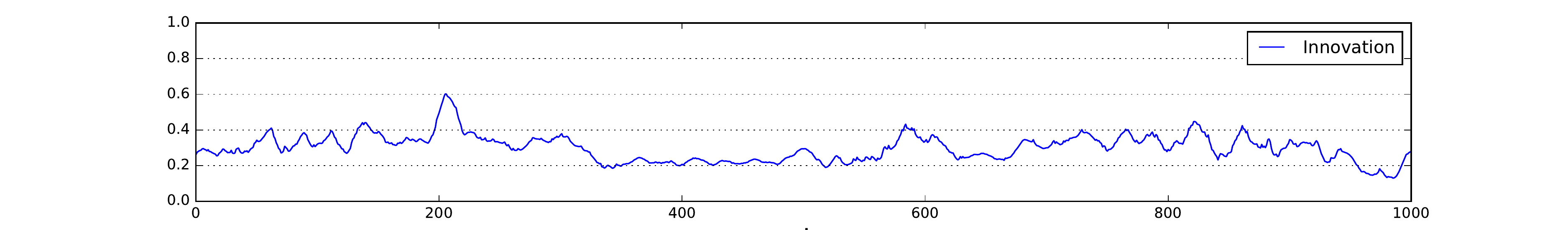}}
\centerline{\scriptsize{(e)}\includegraphics[width=0.96\linewidth,trim={4.8cm 0 3.5cm 0},clip]{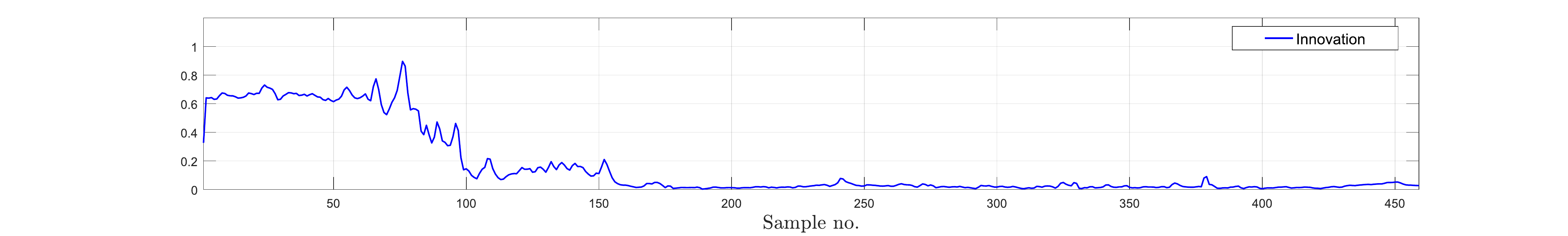}}
\caption{Abnormality in the U-turn avoiding scenario: (a) ground truth labels. (b) and (c) color-coded transition of states $\{C_k\}$ and $\{S_k\}$, respectively. (d) and (e) generated abnormality signal (innovation) from  PL and SL, respectively. The horizontal axis represents the sample number, and the vertical axis shows the innovation values (abnormality signal).}
\label{fig:testushape}
\end{figure}

\noindent{\textbf{Avoiding a pedestrian by a U-turn action}:} In this scenario, which is illustrated in Fig.~\ref{fig:plans}-b, the vehicle performs an avoidance maneuver over a static pedestrian by a U-turn to continue the standard monitoring afterward. The goal is to detect the abnormality, which is the presence of the pedestrian and consequently the unexpected action of the agent with respect to the learned normality during the perimeter monitoring. In Fig. \ref{fig:testushape} the result of anomaly detection from PL and SL representations is shown. The results are related to the highlighted time slice of the testing scenario II (Fig. \ref{fig:subplans}-b).

In Fig. \ref{fig:testushape}-a the green background means that the vehicle moves on a straight line, the blue bars indicate curving, and red show the presence of an abnormal situation (which in this case is the static pedestrian). The abnormality area starts on first sight of the pedestrian and it continues until the avoiding maneuver finishes (end of U-turn). Similarly, the sequences of states $\{C_k\}$ and $\{S_k\}$ in Fig. \ref{fig:testushape}-(b,c), follow the same pattern. While the situation is normal, the super-states repeat the expected normal pattern, but as soon as the abnormality begins the super-state patterns are changed in the both PL and SL (e.g., dummy super states). Furthermore, in the abnormal supper-states the abnormality signals showing the higher values.  
The abnormality signal generated by SL is shown in Fig. \ref{fig:testushape}-e, which represents the innovation between prediction and the observation. The abnormality produced by the vehicle is higher while it moving thought the path which is indicated by red arrows in Fig. \ref{fig:subplans}-b. This is due to the fact that the observations are outside the domain that superstates are trained on. Namely, during the training such state space configuration is never observed. This means the KF innovation becomes higher in the same time interval due to the opposite velocity compared with the normal behavior of the model, more specific, KF innovation is high due to the difference between prediction (which is predicted higher probability going straight) and the likelihood of the observed behavior in a curved path.

\begin{figure}[t]
\centerline{\scriptsize{(a)}\includegraphics[width=0.95\linewidth,height=2cm]{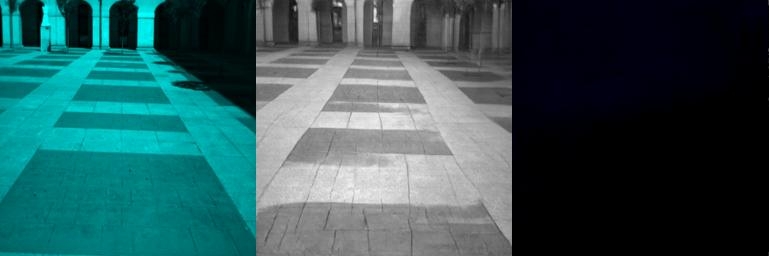}}
\centerline{\scriptsize{(b)}\includegraphics[width=0.95\linewidth,height=2cm]{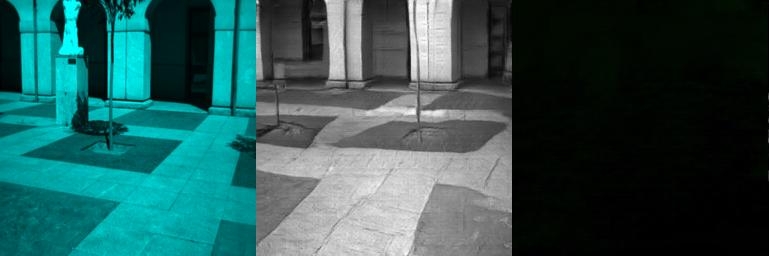}}
\centerline{\scriptsize{(c)}\includegraphics[width=0.95\linewidth,height=2cm]{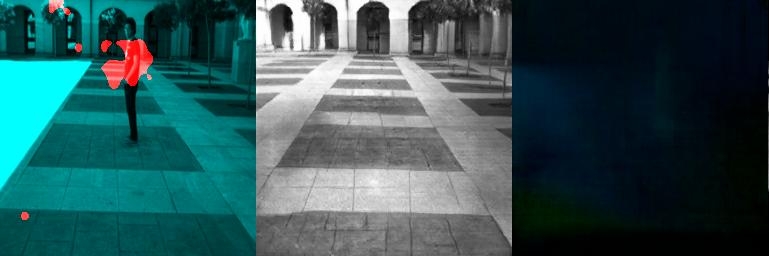}}
\centerline{\scriptsize{(d)}\includegraphics[width=0.95\linewidth,height=2cm]{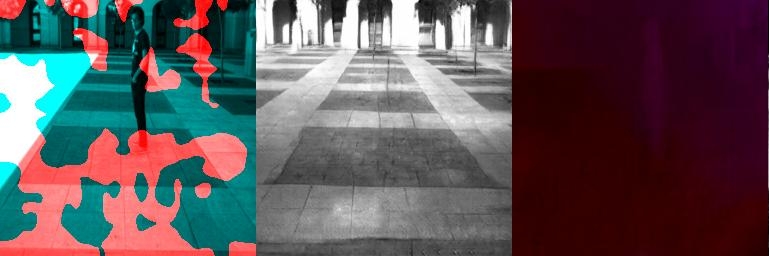}}
\centerline{\scriptsize{(e)}\includegraphics[width=0.95\linewidth,height=2cm]{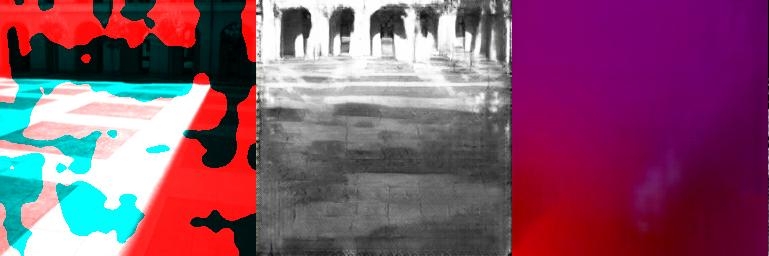}}
\caption{Visualization of abnormality: first column shows the localization over the original frame, second column is the predicted frame, and the last column shows the pixel-by-pixel distance over the optical-flow maps. (a) moving straight, (b) curving, (c) first observation of the pedestrian, (d) and (e) performing the avoiding action.}
\label{fig:visavoiding}
\end{figure}

The abnormality signal generated by PL, Fig. \ref{fig:testushape}-c, is computed by averaging over the distance maps between the prediction and the observation score maps: when an abnormality begins this value does not undergo large changes since the observing a compressed local abnormality (see Fig. \ref{fig:visavoiding}-c) can not change the average value significantly. However, as soon as observing a full sight of pedestrian and starting the avoidance action by the vehicle, the abnormality signal becomes higher since both observed appearance and action are presenting an unseen situation. This situation is shown in Fig. \ref{fig:visavoiding}-(d,e). As soon as the agent back to the known situation (e.g., curving) the abnormality signal becomes lower.

\begin{figure}[t]
\centerline{\scriptsize{(a)}\includegraphics[width=0.96\linewidth,trim={4.35cm 0.6cm 3.2cm 0},clip]{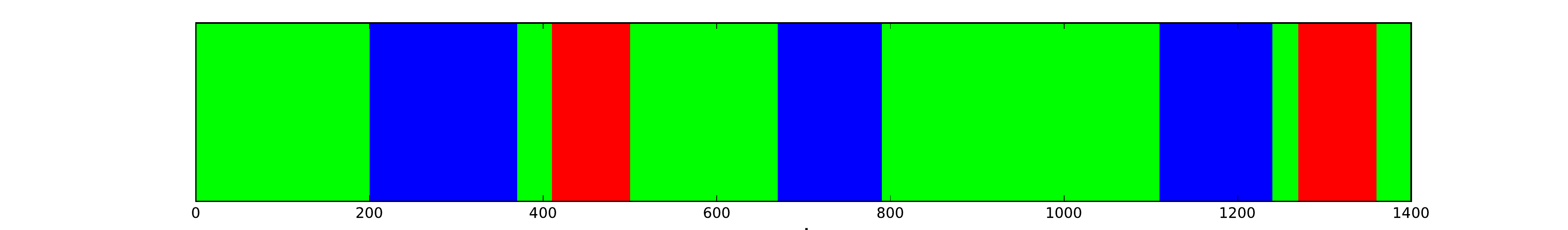}}
\centerline{\scriptsize{(b)}\includegraphics[width=0.96\linewidth,trim={4.35cm 0 3.2cm 0},clip]{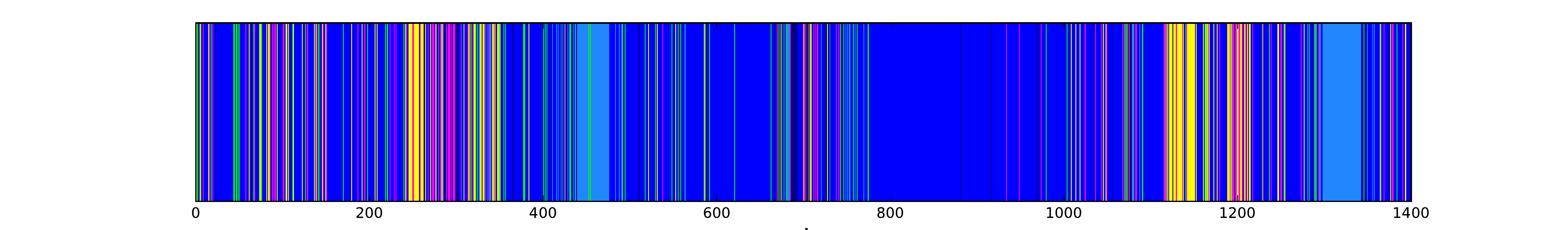}}
\centerline{\scriptsize{(c) }\includegraphics[width=0.94\linewidth,trim={6cm 0 3.8cm 0},clip]{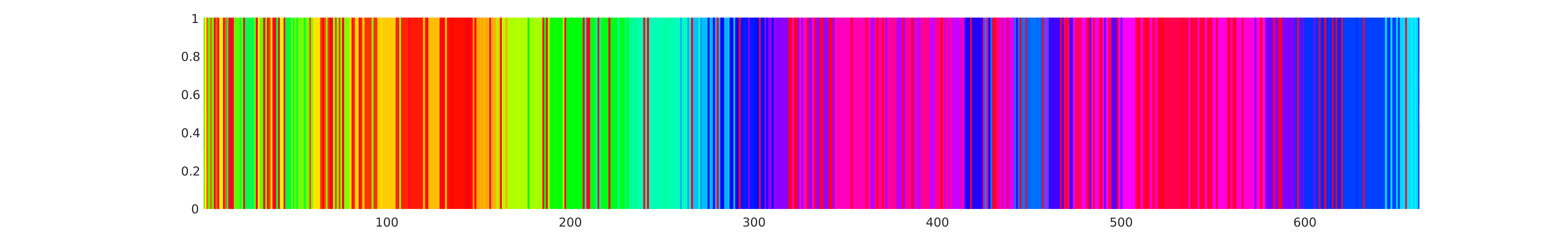}}
\centerline{\scriptsize{(d)}\includegraphics[width=0.96\linewidth,trim={4cm 0 3.2cm 0},clip]{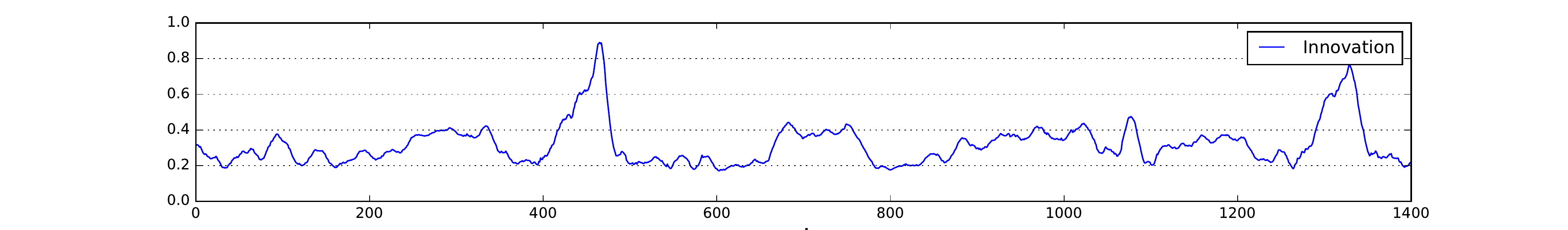}}
\centerline{\scriptsize{(e)}\includegraphics[width=0.96\linewidth,trim={4.8cm 0 3.5cm 0},clip]{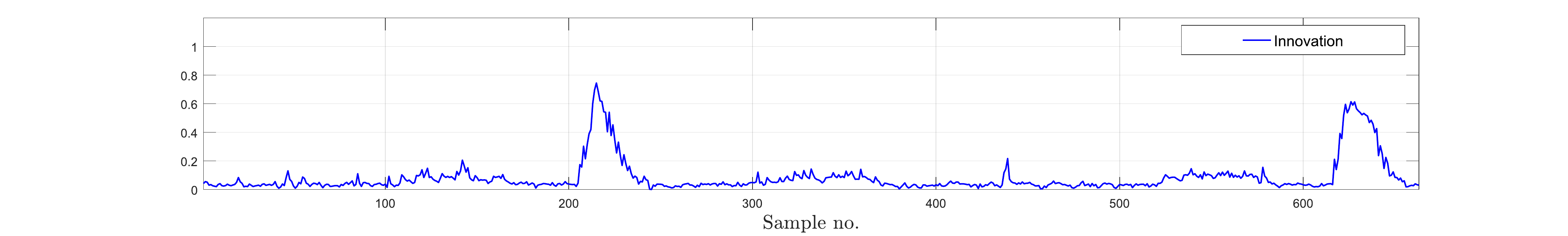}}
\caption{Abnormality in the emergency stop scenario: (a) ground truth labels. (b) and (c) color-coded transition of states from PL and SL, respectively. (d) and (e) generated abnormality signal (innovation) $\{C_k\}$ and $\{S_k\}$, respectively. The horizontal axis represents the sample number, and the vertical axis shows the innovation values (abnormality signal).}
\label{fig:teststop}
\end{figure}
\noindent{\textbf{Emergency stop maneuver}:} This scenario is shown in Fig. \ref{fig:plans}-c, where the agent performs an emergency stop for a pedestrian to cross. Accordingly, the results of abnormality detection, for the highlighted time slice in Fig. \ref{fig:subplans}-c, are represented in Fig. \ref{fig:teststop}.   
In Fig. \ref{fig:teststop}-a the red bars indicate the abnormality areas, where the agent is stopped and waits until the pedestrian cross. Accordingly, this areas are represented as the dummy super states from PL (light-blue color in Fig. \ref{fig:teststop}-b) with high scores in the abnormality signal (see Fig. \ref{fig:teststop}-d). The generated abnormality signal from PL increases smoothly as the agent get better visual to the pedestrian, and reaches to the peak when the agent stops and having a full visual of the pedestrian. Once the pedestrian passes and the agent starts to continue its straight path, the signal drops sharply. Similarly, the abnormality signal from SL representation model (see Fig. \ref{fig:teststop}-e) shows two peaks that correspond to high innovation. Those peaks represent the abnormality patterns associated with the emergency stop maneuver. In contrast with the PL signal, the SL signal reaches to the peak sharply, and then smoothly back to the normal level. Such pattern indicates that the vehicle stops immediately and waits for a while, then it starts again to move and increases its velocity constantly to continue the straight path under the normal situations. As a consequence different motion patterns with respect to those predicted are detected by means of innovation. This is also confirmed by the color-coded super states in Fig. \ref{fig:teststop}-c, where the green and dark-blue states are continued longer than what expected with respect to the normal pattern that learned from the previous observations.

%% file: sections/discussion.tex
\noindent{\textbf{The cross-modal representation}: }One of our novelties in this paper is using GAN for a {\em multi-channel} data representation. Specifically, we use both appearance and motion (optical-flow) information: a two-channel approach which has been proved to be empirically important in previous works. Moreover, we propose to use a cross-channel approach where, we train two networks which respectively transform raw-pixel images in optical-flow representations and vice-versa. The rationale behind this is that the architecture of our conditional generators $G$ is based on an encoder-decoder (see Sec.~\ref{subsec:PL}) and one of the advantage of such channel-transformation tasks is to prevent $G$ learns a trivial identity function and force $G$ and $D$ to construct sufficiently informative internal representations.

\noindent{\textbf{Private layer and shared layer cross-correlation}: }The PL and SL levels are providing complimentary information regarding the situation awareness. As an instance, it has been observed that in case of PL the super-states are invariant to the agent's location, while SL super-states representation is sensitive to such spatial information. In other words, PL representation can be seen as the semantic aspect of agent's situation awareness (e.g., moving straight, curving) regardless to the current location of the agent. That's why the pattern of super-states sequences is repeated with respect to the agent's taken action (see Fig. \ref{fig:pri_states}-b). However, the SL representation is included the spatial information the sequence of super-states could be different with respect to the agent's location (see Fig. \ref{fig:pri_states}-c). This issue is more obvious in the example of U-shape avoiding experiment (see Fig. \ref{fig:testushape}). In this case after performing avoiding maneuver the abnormality signal in the PL back to the normal as well as the super-states sequences, while the abnormality signal from SL remains high due to this space position dependency. In light of the above, these two representations are carrying complimentary information and finding a cross-correlation between PL and SL situation representation (e.g., using coupled Bayesian network) could potentially increase the ability of anomaly detection and consequently boost the entire self-awareness model.

\noindent{\textbf{Modalities alignment}: }In our work, the means of alignment between two modalities localization (in SL) and visual perception (in PL), is provided by the synchronous time stamps assigned from the sensors. Since the time reference is equal for both sensors (odometer and camera) this is possible to collect the aligned multi-modal data. However, another advantage of modeling the cross-correlation between two layers (PL and SL) could be providing an extra reference for finding such alignments. Namely, the cross-correlation of repetitive patterns in PL and SL can be used for the data alignment. In case of having asynchronous sensors, this could be useful to find the exact alignment.